\documentclass{ecai} 

\usepackage{latexsym}
\usepackage{amssymb}
\usepackage{amsmath}
\usepackage{amsthm}
\usepackage{booktabs}
\usepackage{enumitem}
\usepackage{graphicx}
\usepackage{color}
\usepackage{bm}
\usepackage{multirow}
\usepackage{pifont}
\usepackage{xspace}
\usepackage{tabularx}
\usepackage{hyperref}
\hypersetup{hidelinks, colorlinks=true, allcolors=black, pdfstartview=Fit, breaklinks=true}
 
\newcommand{\BibTeX}{B\kern-.05em{\sc i\kern-.025em b}\kern-.08em\TeX}

\begin{document}


\begin{frontmatter}


\title{Less is More: Efficient Brain-Inspired Learning for Autonomous Driving Trajectory Prediction}

\author[1]{\fnms{Haicheng}~\snm{Liao}}
\author[2]{\fnms{Yongkang}~\snm{Li}}
\author[1]{\fnms{Zhenning}~\snm{Li}}
\author[1]{\fnms{Chengyue}~\snm{Wang}}
\author[1]{\fnms{Yuming}~\snm{Huang}}
\author[1]{\fnms{Chunlin}~\snm{Tian}}
\author[3]{\fnms{Zilin}~\snm{Bian}}
\author[1]{\fnms{Kaiqun}~\snm{Zhu}}
\author[4]{\fnms{Guofa}~\snm{Li}}
\author[5]{\fnms{Jia}~\snm{Hu}}
\author[6]{\fnms{Ziyuan}~\snm{Pu}}
\author[7]{\fnms{Zhiyong}~\snm{Cui}}
\author[1]{\fnms{Chengzhong}~\snm{Xu}}
\address[1]{University of Macau} \address[2]{UESTC} \address[3]{New York University} \address[4]{Chongqing University} \address[5]{Tongji University} \address[6]{Southeast University} \address[7]{Beihang University}


\begin{abstract}
Accurately and safely predicting the trajectories of surrounding vehicles is essential for fully realizing autonomous driving (AD). This paper presents the Human-Like Trajectory Prediction model (HLTP++), which emulates human cognitive processes to improve trajectory prediction in AD. HLTP++ incorporates a novel teacher-student knowledge distillation framework. The ``teacher'' model equipped with an adaptive visual sector, mimics the dynamic allocation of attention human drivers exhibit based on factors like spatial orientation, proximity, and driving speed. On the other hand, the ``student'' model focuses on real-time interaction and human decision-making, drawing parallels to the human memory storage mechanism. Furthermore, we improve the model's efficiency by introducing a new Fourier Adaptive Spike Neural Network (FA-SNN), allowing for faster and more precise predictions with fewer parameters. Evaluated using the NGSIM, HighD, and MoCAD benchmarks, HLTP++ demonstrates superior performance compared to existing models, which reduces the predicted trajectory error with over 11\% on the NGSIM dataset and 25\% on the HighD datasets. Moreover, HLTP++ demonstrates strong adaptability in challenging environments with incomplete input data. 
This marks a significant stride in the journey towards fully AD systems.
\end{abstract}
\end{frontmatter}

\section{Introduction}
As we stand on the brink of a revolution in autonomous vehicles (AVs), the essential challenge that emerges is not just in the engineering of these vehicles but in endowing them with a cognitive prowess akin to human drivers. Central to this journey is the challenge of trajectory prediction, a task that requires a nuanced understanding of both the external driving environment and the internal cognitive processes of decision-making \cite{huang2022survey}. Traditional deep-learning models in AVs have made significant strides in data processing and pattern recognition but still encountered a bottleneck in complex and data-missing scenery. Given that human drivers are able to navigate complex scenes with remarkable adaptability and foresight, we aim to emulate some of the decision-making processes of human drivers to address this issue.
\begin{figure}[t]
    \centering
    \includegraphics[width=\linewidth]{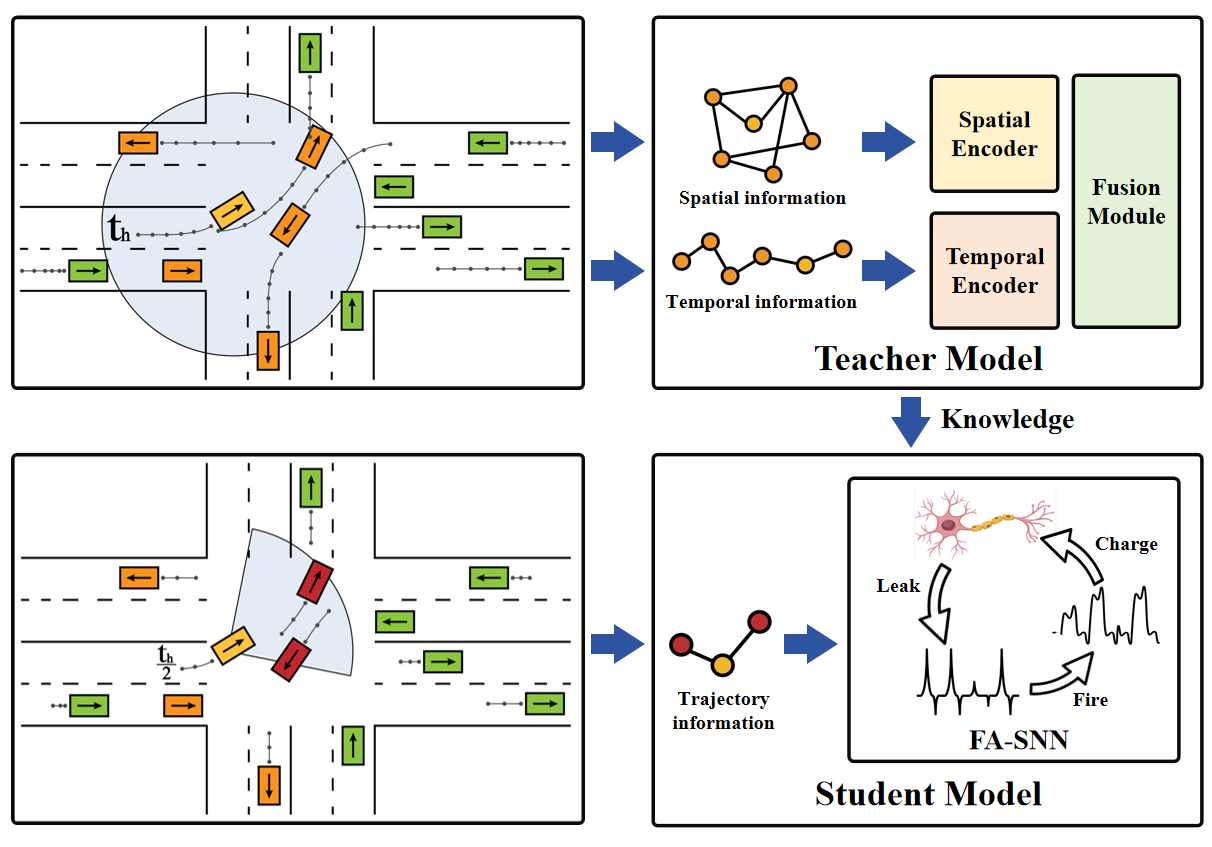}
    \caption{Illustration of our proposed model HLTP++. The ``teacher'' model is designed to imitate the attention distribution in human driving by capturing contextual data such as surrounding agents and temporal-spatial interactions, generating prior knowledge. The ``student'' model, guided by this knowledge and leveraging FA-SNN, emulates neural transmission and provides efficient and human-like trajectory prediction with only 50\% of the observations provided to the ``teacher'' model.}
    \vspace{2em}
    \label{fig:1}
\end{figure}
At the heart of human decision-making process in driving lies a sophisticated network of neural activities, which involves multiple regions of the brain, each playing a distinct yet interconnected role \cite{louie2018working}. The occipital and temporal lobes, responsible for visual processing, work in concert with the prefrontal and parietal cortex, which are central to decision-making and spatial reasoning \cite{miller2016key}. It is this harmonious interplay that allows human drivers to anticipate potential hazards and make informed decisions based on a blend of current sensory input and past experience \cite{baddeley2013essentials}.

Drawing inspiration from these cognitive mechanisms, our Human-Like Trajectory Prediction (HLTP++) model aims to replicate the essence of human decision-making. Due to the dual advantages of being lightweight and capable of applying cognitive mechanisms, we utilize knowledge distillation as the overarching framework. As shown in Figure \ref{fig:1}, the ``teacher" model of HLTP++ is designed to mimic the brain's visual processing faculties, employing a neural network architecture with visual pooling to prioritizes and interprets incoming visual data. This aspect of the model is analogous to the way the occipital and temporal lobes filter, process, and relay visual information to other parts of the brain. The ``student" model of HLTP++, in turn, introduces a novel Spike Neural Network FA-SNN that represents the decision-making and reasoning aspects of the prefrontal and parietal cortex. Our goal is to create a model that not only processes information as the human brain does but also adapts to new information faster with similar flexibility and accuracy, aiming to address the current issues of high parameter counts  and unhuman-like predicted trajectories.
Overall, the contributions of HLTP++ are multifaceted:
 \begin{itemize}
    \item  We introduce a novel visual pooling mechanism to emulate the dynamic visual sector of human observation and adaptively self-adjust its attention to different agents in central and peripheral vision in real time under different scenes. Additionally, the Fourier Adaptive Spike Neural Network (FA-SNN) is proposed to handle traffic scenes with missing data by mimicking the neuronal pulse propagation process in human brain.
    
    \item The HLTP++ presents a heterogeneous teacher-student knowledge distillation framework by using Knowledge Distillation Modulation (KDM) for multi-level tasks. This approach automatically adjusts the ratio between loss functions, significantly facilitating model training in complex trajectory distillation situations.

    \item Benchmark tests on the NGSIM, HighD, and MoCAD datasets have shown that the HLTP++ outperforms existing top baselines by considerable margins, demonstrating its superior robustness and accuracy in various traffic conditions, including highways and dense urban environments. Notably, it exhibits remarkable performance even with fewer input observations and in scenarios characterized by missing data.
\end{itemize}

\section{Related Work}\label{Related Work}
\textbf{Trajectory Prediction.} In the realm of AVs, deep learning has catalyzed the adoption of various neural network architectures. Prior studies have explored Recurrent Neural Networks (RNNs) \cite{quan2021holistic}, social pooling \cite{zhang2020multi}, Graph Neural Networks (GNNs) \cite{li2019grip, hltp2024liao}, attention mechanisms \cite{zhou2022hivt,liao2024cognitive}, and Transformers framework \cite{li2023context,zhou2023query,liao2024characterized} to enhance spatial and temporal feature extraction from trajectory data. Concurrently, some studies \cite{kamenev2022predictionnet,Cui_2021} focus on improving safety in trajectory predictions, while others \cite{hu2023planningoriented,liao2024gpt} develop end-to-end models incorporating multimodality. 
 
However, existing models' computational intensity and lack of human-centric driving cognition pose challenges for real-time application and safety. To address this, we introduce a lightweight yet robust model that mimics human driving habits, enhancing safety for real-time trajectory prediction.\\
\textbf{Knowledge Distillation}. Knowledge distillation, originally proposed by Hinton et al. \cite{hinton2015distilling}, involves a complex "teacher" model transferring knowledge to a simpler "student" model. Initially aimed at model compression, its applications have expanded to improve generalization \cite{zagoruyko2016paying} and increase accuracy \cite{porrello2020robust}, while reducing computational and time requirements \cite{bhardwaj2019efficient,liu2023hierarchical}. However, there have been only a few applications in the autonomous driving field \cite{hltp2024liao}. 
This study integrates human cognitive processes into the trajectory prediction paradigm by developing knowledge distillation method. The ``teacher'' model emulates human reasoning and effectively transfers this knowledge to the ``student'' model, simulating human driving behavior. In addition, we propose a novel multi-level task learning paradigm in the loss functions of our model for efficient training.\\
\textbf{Spike Neural Network}.  In recent years, there has been an increasing demand for edge computing applications. To address this, third-generation neural networks have been proposed. The most prominent representative among them is the SNN. Itis influenced by the neurons of the biological nervous system, which use discrete spikes for information encoding and transmission \cite{Diehl2015UnsupervisedLO}. Research has shown the effectiveness of SNNs \cite{ijcai2023p342}, particularly in terms of energy-saving, making them suitable for embedded deployment in vehicular systems. Traditionally, the primary focus of SNN studies has been on reducing inference time, often at the expense of overlooking improvements in adaptability. It has been observed that existing SNNs lack adaptability and temporal feature extraction, primarily due to fixed threshold constraints. Hence, this study introduces FA-SNN to improve the adaptability and temporal analysis of the trajectory prediction model.

\begin{figure}[t]
  \centering
  \includegraphics[width=0.75\linewidth]{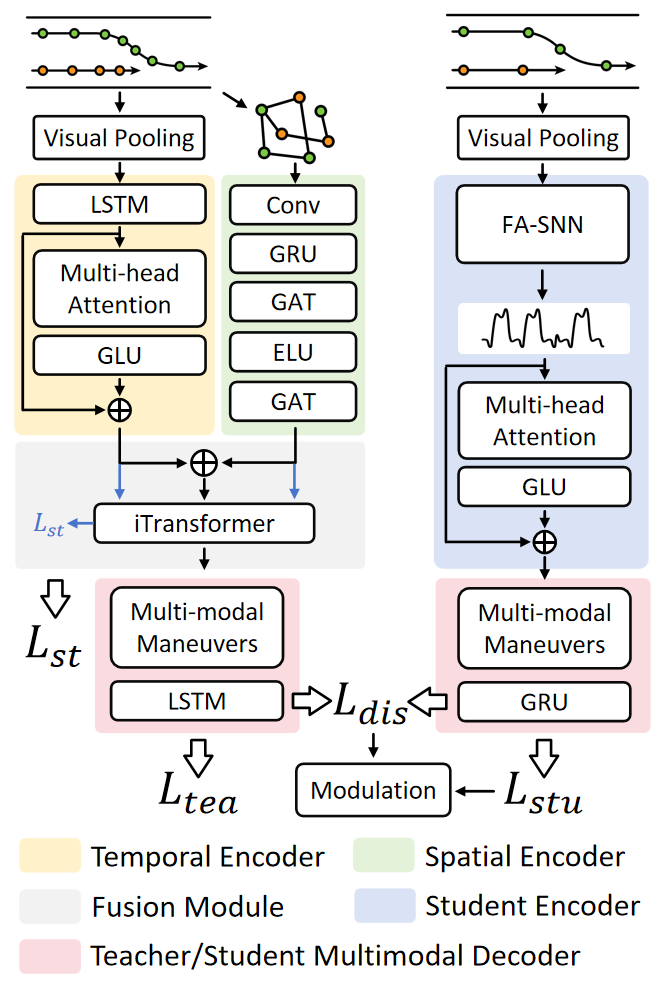}
  \caption{Overall ``teacher-student'' architecture of the HLTP++. The ``teacher'' model employs temporal and spatial encoders to capture spatio-temporal interactions, followed by a fusion module that integrates features. These fusion features are then fed into the decoder to produce multimodal predictions based on various potential maneuvers. The ``student'' model acquires knowledge from the ``teacher'' model through the KDM training strategy and incorporates FA-SNN for efficient, human-like trajectory prediction.}
  \label{fig1}
  \vspace{2em}
\end{figure}

\section{Methodology}\label{Method}
\subsection{Problem Formulation}
The principal objective of this study is to predict the trajectory of the target vehicle, encompassing all traffic agents within the sensing range of the AV in a mixed autonomy environment where AVs coexist with human-driven vehicles. At any time $t$, the state of the $i$th traffic agents can be denoted as $p^{i}_{t}$, where $p^{i}_{t}=(x^{i}_{t}, y^{i}_{t})$ represents the 2D trajectory coordinate. Given the trajectory coordinate data of the target vehicle (superscript 0) and all its observed traffic agents (superscripts from 1 to $n$) in a traffic scene in the interval $[1, T_{obs}]$, denoted as $\bm{X}=\{p^{0: n}_{t}\}_{t=1}^{T_{obs}} \in \mathbb{R}^{(n+1) \times T_{obs} \times 2}$, the model aims to predict a probabilistic multi-modal distribution over the future trajectory of the target vehicle, expressed as $P(\bm{Y}|\bm{X})$. Here, $\bm{Y}=\{\bm{y}^{0}_{t}\}_{t=T_{obs}+1}^{T_{f}} \in \mathbb{R}^{T_{f} \times 2}$ is the predictive future trajectory coordinates of the target vehicle over a time horizon $T_{f}$, and $\bm{y}^{0}_{t}=\{(\tilde{p}^{0,1}_{t}; \tilde{c}^{0,1}_{t}),
(\tilde{p}^{0,2}_{t};\tilde{c}^{0,2}_{t}),\cdots, (\tilde{p}^{0,C}_{t};\tilde{c}^{0,C}_{t})\}$ encompasses both the potential trajectory and its associated maneuver likelihood ($\Sigma^{1}_{\bm{C}} {c}^{0,i}_{t}=1$), with $\bm{C}$ denoting the total number of potential trajectories predicted. Notably, we focus on optimising the output of the ``student'' model, and the predictive future trajectory coordinates  $\bm{Y}$ in this study is the output of the ``student'' model, formulated as $\bm{Y}_t^{stu}=\bm{Y}$. 

\subsection{Scene Representation}

HLTP++ describes scenarios by focusing on the relative positions of traffic agents, aligning with human spatial understanding. It preprocesses historical data into two key spatial forms: 1) visual vectors $\mathcal{S}$ that capture relative position, velocity, and acceleration $\mathcal{S}=\{\mathcal{S}_{\Delta p}, \mathcal{S}_{\Delta s}, \mathcal{S}_{\Delta a}\} \in \mathbb{R}^{(n+1) \times T_{obs} \times 4}$ of the target vehicle relative to its neighbors; 2) context matrices $\mathcal{M}$ describing speed and direction angle differences $\mathcal{M}=\{\mathcal{M}_{\Delta s}, \mathcal{M}_{\Delta \theta}\} \in \mathbb{R}^{(n+1) \times T_{obs} \times 2}$ among the surrounding agents. 

\subsection{Overall Architecture}
Figure \ref{fig1} illustrates the architecture of HLTP++. We use a novel pooling mechanism with an adaptive visual sector for data preprocessing. This sector dynamically adapts to capture important cues in different traffic situations which is similar to attention allocation. Additionally, the model also employs a teacher-student heterogeneous network distillation approach for human-like trajectory prediction.
i) \textbf{The ``teacher'' model}. The Temporal Encoder and the Spatial Encoder within the ``teacher'' model process visual vectors and context matrices to produce temporal and spatial feature vectors, respectively. These vectors are then fed into the Fusion Module to fuse two modalities. The output of the Fusion Module is then fed into the Teacher Multimodal Decoder, which enables the prediction of different potential maneuvers for the target vehicle, each with associated probabilities.
ii) \textbf{The ``student'' model}. The Fourier Adaptive SNN first processes trajectory temporal information from the Visual Pooling by imitating the transmission of neurons. Then the output matrix is fed into the Student Multimodal Decoder which is similar to the teacher. Besides self-training, the ``student'' model acquires knowledge from the ``teacher'' model using a Knowledge Distillation Modulation (KDM) training strategy. This approach ensures accurate trajectory predictions while requiring fewer input observations.

\subsection{Visual Pooling Mechanism}
Research \cite{tucker2021speeding} shows that human drivers, constrained by brain's working memory, focus mainly on a few external agents in their central visual field, especially in high-risk situations. The driver's visual sector, influenced by speed, narrows at higher speeds for focused attention and widens at lower speeds for broader awareness.

HLTP++ introduces a visual pooling mechanism that emulates this adaptive visual attention. It features an adaptive visual sector that adjusts the field of view based on vehicle speed. Specifically, in contrast to models with uniform attention distribution, we propose a visual weight matrix $H$ that adapts to changing focus in the central visual field at different speeds. Speed thresholds at $0 km/h$, $30 km/h$, $60 km/h$, and $90 km/h$ define distinct values of the visual sectors. This approach refines the understanding of attention during driving. The visual weight matrix $H_{\textit{vision}}$ is then integrated by the input visual vectors $\mathcal{S}$. Formally,
\begin{equation}
\tilde{\mathcal{S}} = H_{\textit{vision}} \odot \mathcal{S},
\end{equation}
This equation produces visual vectors $\tilde{\mathcal{S}}$ that encapsulate human drivers' varying attention patterns.

\subsection{Teacher Model}
The ``teacher"  model integrates Temporal and Spacial encoders to closely emulate human visual perception, mirroring the retinal processing of the human drivers. As an enhancement, it further employs iTransformer framework in the decoder to effectively extract spatio-temporal interactions. \\
\textbf{Temporal Encoder.} In real-world driving, the human brain, with its limited processing capacity, prioritizes information to facilitate efficient decision-making. 
Therefore, allocating distinct attention to different features is essential, as it reduces the cognitive load on the brain in processing non-essential information. We employ a LSTM layer to process temporal information, followed by a multi-head attention mechanism for attention allocation. \\
\textbf{Spatial Encoder.} Human drivers focus on their central vision while continuously monitoring their peripheral vision through side and rear-view mirrors to understand their surroundings, including nearby vehicles, pedestrians, and road conditions. To replicate this peripheral monitoring, especially during maneuvers, we introduce the Spatial Encoder. It processes a quarter of the time-segmented matrices $\mathcal{M}  \in \mathbb{R}^{(n+1) \times \frac{T_{obs}}{4} \times 2}$ with a $1\times 1$ convolutional layer for channel expansion, followed by a $3\times 3$ layer for specific feature extraction, incorporating batch normalization and dropout for robustness. Enhanced with Graph Attention Networks and ELU activation, it produces spatial vectors $\mathbf{O}_{\textit{s}}$.\\
\textbf{Fusion Module.} The combined outputs of the Spatial Encoder $\mathbf{O}_{\textit{s}}$ and the Temporal Encoder $\mathbf{O}_{\textit{t}}$ are fused and then fed into the iTransformer architecture \cite{liu2023itransformer} for advanced spatio-temporal interaction analysis, generating hidden states $\mathbf{I}$. Furthermore, we use the disparities between temporal and spatial features to generate a loss, denoted as $L_{st}$, which serves as one of the loss functions for training the ``teacher'' model.\\
\textbf{Teacher Multimodal Decoder.} 
The decoder of the ``teacher" model, based on a Gaussian Mixture Model (GMM), accounts for uncertainty in trajectory prediction by evaluating multiple possible maneuvers and their probabilities. Specifically, built on the base layer of estimated maneuvers $\bm{C}$, the model assumes that the probability distribution for trajectory predictions follows a Gaussian framework:
\begin{equation}
    P_{\bm{\Omega}} (\bm{Y}|\bm{C},\bm{X}) = N(\bm{Y}|\mu(\bm{X}),\Sigma(\bm{X}))
\end{equation}
where $\bm{X}$ represents the input to our model, and $\bm{\Omega}=\left[\Omega^{t+1}, \ldots,\Omega^{t+t_{f}}\right]$ symbolizes the estimable parameters of the distribution. Each $\Omega^{t}=[\mu^{t},\Sigma^{t}]$ represents the mean and variance for the predicted trajectory at time $t$. Then, in the next layer, the multimodal predictions are conceptualized as a GMM:
\begin{equation}\label{eq.6}
    P(\bm{Y}|\bm{X})=\sum_{\forall i} P\left(c_{i}|\bm{X}\right) P_{\bm{\Omega}}\left(\bm{Y}|c_{i},\bm{X}\right)
\end{equation}
where $c_{i}$ denotes the $i$-th maneuver in $\bm{C}$. 

Then, the hidden states are processed through softmax activation and a MLP layer, forming a probability distribution $\mathbf{O}_{\textit{tea}}$ over potential future trajectories. This approach balances the precision and validity, which is critical for dynamic and uncertain driving environments.

\subsection{Student Model}
The FA-SNN is introduced in the ``student'' model, which emphasizes the short-term and fewer observations, with visual vectors $\tilde{\bm{S}}$ ($T_{obs} =8$), utilizing visual vectors and a lightweight architectural design for efficient learning. In a departure from the complex framework of the ``teacher'' model, the ``student'' model employs a more lightweight architecture to produce a multimodal prediction distribution $\mathbf{O}_{stu}$. By learning the behavioral paradigm from the ``teacher'' model, the ``student'' model is able to make human-like predictions even when constrained by limited observations.
\begin{figure}[t]
    \centering
\includegraphics[width=0.75\linewidth]{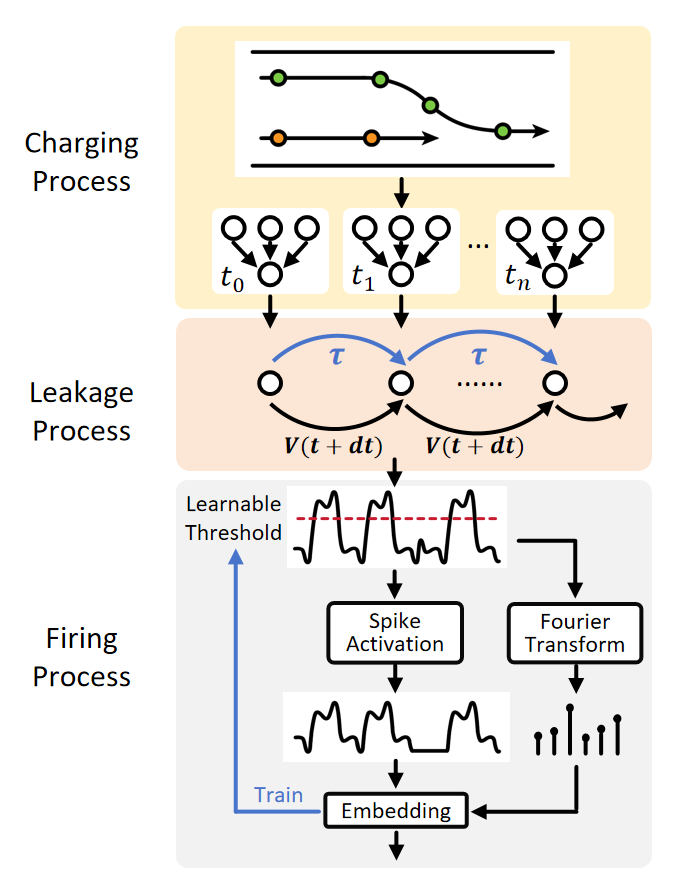}
    \caption{Overall architecture of the FA-SNN.}
    \vspace{2em}
    \label{fig:snn}
\end{figure}

Specifically, The FA-SNN proposed in this study is an enhanced version of the traditional SNN model. It addresses the challenge of fitting difficulties during training by introducing adaptive adjustments and optimizing temporal feature extraction. The approach is based on the idea that neurons in an SNN should adapt to different scenarios, which is mainly reflected in the adjustment of the threshold magnitude. Inspired by the Leaky Integrate-and-Fire (LIF) mechanism, the Fourier Transform (FT) is used to extract specific features. More specifically, the FA-SNN's forward propagation involves three essential processes: Charging, Leakage, and Firing Processes.\\
\textbf{Charging Process.} Similar to traditional perceptron neurons, the current neuron is charged by aggregating input spike sequences from previous neurons through varying weights at discrete time steps $\{t_0, t_1, ..., t_n, ..., t_N\}$.\\
\textbf{Leakage Process}. In the LIF mechanism, neurons experience leakage due to voltage differences in their surroundings. The internal voltage $V$ of the spiking neuron tends towards an equilibrium voltage $U$ over time $t$, adhering to the differential equation $U-V=-\eta\frac{dV}{dt}$, where $\eta=1$ denote the leakage decay rate, indicating the magnitude of voltage decay. This dynamic is pivotal for the neuron's voltage stabilization, allowing the calculation of future voltage states. After solving the above equation, we can obtain:
\begin{equation}
V(t_n)=U-Ce^{-\frac{t_n}{\eta}}
\end{equation}
where $C$ is a constant value. This allows us to calculate the voltage at the next moment $t_{n+1}$:
\begin{equation}
\begin{aligned}
    V(t_{n+1})=V(t_n+dt)=e^{-\frac{dt}{\eta}}(V(t)-U)+U
\end{aligned}
\end{equation}
\textbf{Firing Process.} The firing process is activated based on the spike magnitude through an activation function. Given a spike threshold $U_0$, the voltage $V'(t_{n+1})$ can be defined as follows:
\begin{equation}
V'(t_{n+1}) =
\begin{cases}
V(t_{n+1})-U_0, & V(t_{n+1})>U_0 \\
V(t_{n+1}),   & V(t_{n+1})\leq U_0 \\
\end{cases}
\end{equation}
Unlike traditional models with a fixed spike threshold, the proposed FA-SNN employs a learnable threshold, adjusting dynamically to preserve temporal features. Importantly, a Fast Fourier Transform (FFT) is applied to the pre-firing spike value $V(t_{n+1})$ to incorporate frequency information, enhancing the model's representation capability:
\begin{equation}
F[V(t_{n+1})] = \sum_{\tau=0}^{N} V(t_{n+1}) \cdot e^{-\frac{2\pi i}{N+1} (n+1)\tau}
\end{equation}
where $i$ is the imaginary unit. According to Euler's formula: 
\begin{equation}
e^{-\frac{2\pi i}{N+1} (n+1)\tau}=cos[-\frac{2\pi(n+1)}{N+1}\tau]+isin[-\frac{2\pi(n+1)}{N+1}\tau],
\end{equation}
where $A=cos[-\frac{2\pi(n+1)}{N+1}\tau]$ and $B=sin[-\frac{2\pi(n+1)}{N+1}\tau]$ respectively represent the real and imaginary parts of $e^{-\frac{2\pi i}{N+1} (n+1)\tau}$. We then compute the power spectrum $W=(|A|+|B|)^2$ to serve as the output feature, where ``$||$'' represent the absolute value symbol.
\\

\textbf{Backpropagation.} Due to the discontinuity of the activation function used during SNN firing, conventional chain-rule differentiation is infeasible. To circumvent this, the gradient $G$ is redefined, factoring in the spike threshold and introducing parameters like the absolute width $w_a$, gradient width $w_g$ and gradient scale $s$:
\begin{equation}
G(V'(t_{n+1})) = \frac{s}{w_a} \times \exp\left(-\frac{|V'(t_{n+1})-U_0|}{w_a}\right)
\end{equation}
where $w_a = U_0 \cdot w_g$, and $w_g = 0.5, \, s = 1.0$.

\section{Training}\label{Training_1}
\subsection{Teacher Training} 
For the ``teacher'' model, we follow a standard protocol, using 3 seconds of observed trajectory for input ($T_\textit{obs}=16$) and predicting a 5-second future trajectory ($T_\textit{f}=25$). To extract complex knowledge from the dataset, we allow slight overfitting during training. The training loss function of the teacher model consists of three parts: trajectory loss $\mathcal{L}^{tea}_{traj}$, maneuver loss $\mathcal{L}^{tea}_{man}$ and temporal-spacial loss $L_{st}$ from iTransformer.\\
\textbf{Trajectory Loss.} In our trajectory prediction process, we treat the output 2D trajectory coordinates $P=(x,y)$ as a bivariate Gaussian distribution. Therefore, we could use Negative Log-Likelihood (NLL) loss $\mathcal{L}^{tea}_{traj}$ to measure the disparity between the prediction and the ground truth. Considering the total number $C$ of potential trajectories predicted, with $P_{pred}^{tea}$ and $P_{gt}$ representing the teacher model's predicted trajectory coordinates and the ground truth coordinates respectively, the trajectory loss $\mathcal{L}^{tea}_{traj}$ for the teacher model can be formulated as: $\mathcal{L}^{tea}_{traj}=\sum_{t}^{T_f}\sum_{c}^{C}\mathcal{L}^{N}(P_{pred}^{tea},P_{gt})$. Mathematically, for a specific data point, given the model's predicted probability distribution $P(\bm{Y}|\bm{X})$, where $\bm{Y}$ represents the output future trajectory and $\bm{X}$ denotes the input features, the NLL Loss $\mathcal{L}^{N}$ is defined as follows:
\begin{equation}
\begin{aligned}
\mathcal{L}^{N}=-log(P(\bm{Y}|\bm{X})),
\end{aligned}
\end{equation}
\textbf{Maneuver Loss.} To address the potentially detrimental effects of mis-classifying maneuver types on trajectory prediction accuracy and robustness, we adopt the Mean Squared Error (MSE) loss $\mathcal{L}^{tea}_{man}$ to measure the disparity between the predicted maneuver types $M_{pred}^{tea}$ and the ground truth maneuver types $M_{gt}^{tea}$, which is formulated as: $\mathcal{L}^{tea}_{man}=\sum_{t}^{T_f}\sum_{c}^{C}\mathcal{L}^{M}(M_{pred}^{tea},M_{gt})$,
so the total loss function of teacher model is formulated as follows:
\begin{equation}
    \mathcal{L}^{tea} = \mathcal{L}^{tea}_{traj} + \mathcal{L}^{tea}_{man} + \mathcal{L}_{st},
\end{equation}

\subsection{Student Training}

The ``student'' model is trained to predict 5-second future trajectories with fewer input observations. To improve its predictive performance with limited observations, we decouple the total loss function of the student model $\mathcal{L}$ as the student loss (especially refers to the loss function exclusively associated with the ``student'' model) $\mathcal{L}^{stu}$ and distillation loss $\mathcal{L}^{dis}$. The student loss, similar to that of the ``teacher'' model, quantifies the discrepancy between the model's predicted trajectories, maneuvers and their ground truths. Formally,
\begin{equation}
\begin{aligned}
    \mathcal{L}^{stu} &= \mathcal{L}^{stu}_{traj} + \mathcal{L}^{stu}_{man} \\&= \sum_{t}^{T_f}\sum_{c}^{C}\left(\mathcal{L}^{N}(P_{pred}^{stu},P_{gt})+\mathcal{L}^{M}(M_{pred}^{stu},M_{gt})\right),
\end{aligned}
\end{equation}
where $P_{pred}^{stu}$ and $M_{pred}^{stu}$ represent the predicted 2D coordinates and maneuvers of the ``student'' model.

Moreover, we apply the MSE loss to measure the disparity between the outputs of the teacher and the ``student'' model:
\begin{equation}
\begin{aligned}
    \mathcal{L}^{dis} &= \mathcal{L}^{dis}_{traj} + \mathcal{L}^{dis}_{man} \\&= \sum_{t}^{T_f}\sum_{c}^{C}\left(\mathcal{L}^{M}(P_{pred}^{stu},P_{pred}^{tea})+\mathcal{L}^{M}(M_{pred}^{stu},M_{pred}^{tea})\right),
\end{aligned}
\end{equation}
Hence, the total loss function of the ``student'' model is formulated as
$\mathcal{L}= \mathcal{L}^{stu}+ \mathcal{L}^{dis}$. Then, we propose a method for tuning multiple tasks that evaluates the importance of different loss functions and automatically adjusts the weights between them for efficient training.

\subsection{Knowledge Distillation Modulation}

Given that $\mathcal{L}$ is composed of sub-loss functions from multiple tasks, determining the proportionality relationships between them poses a challenging problem. 
Both $\mathcal{L}^{stu}$ and $\mathcal{L}^{dis}$ are further decomposed into maneuver loss function $\mathcal{L}_{traj}$ and trajectory coordinate loss function $\mathcal{L}_{man}$:
\begin{equation}
\begin{aligned}
\mathcal{L}^{stu} = \mathcal{L}^{stu}_{traj} + \mathcal{L}^{stu}_{man}, ~~~~~~ 
\mathcal{L}^{dis} = \mathcal{L}^{dis}_{traj} + \mathcal{L}^{dis}_{man},
\end{aligned}
\end{equation}

Drawing the inspiration from the notable work \cite{kendall2018multitask}, we incorporate the KDM to weight the \textit{trajectory} loss and the \textit{distillation} loss, with homoscedastic uncertainty. To the best of our knowledge, we are the first to propose a multi-level, multi-task hyperparameter tuning approach to autonomously adjust knowledge distillation hyperparameters during training in this field. Our approach defines a multi-level task where the overarching training loss function is composed of an ensemble of sub-loss functions. Each of these sub-loss functions is further composed of additional sub-loss functions that share some degree of similarity. 

Following the approach outlined in Kendall et al. \cite{kendall2018multitask} for the first level of the loss function, we obtain the following equation:
\begin{equation}\label{l1}
\begin{aligned}
\mathcal{L}^{stu}=\frac{1}{2 \sigma_{t}^2} \mathcal{L}^{stu}_{traj}(\mathbf{W})+\frac{1}{2 \sigma_{m}^2} \mathcal{L}^{stu}_{man}(\mathbf{W})+ log\sigma_{t}\sigma_{m},\\
\mathcal{L}^{dis}=\frac{1}{2 \sigma_{t}^2} \mathcal{L}^{dis}_{traj}(\mathbf{W})+\frac{1}{2 \sigma_{m}^2} \mathcal{L}^{dis}_{man}(\mathbf{W})+ log\sigma_{t}\sigma_{m},\\
\end{aligned}
\end{equation}
where $\sigma_t, \sigma_m$ are the learnable uncertainty variances, $\mathbf{W}$ represents trainable parameters of the model. Since $\mathcal{L}$ is composed of $\mathcal{L}^{stu}$ and $\mathcal{L}^{dis}$, we use the multi-task tuning approach for the second level:
\begin{equation}\label{l2}
\begin{aligned}
\mathcal{L}=\frac{1}{2 \sigma_{s}^2} \mathcal{L}^{stu}(\mathbf{W})+\frac{1}{2 \sigma_{d}^2} \mathcal{L}^{dis}(\mathbf{W})+ log\sigma_{s}\sigma_{d},
\end{aligned}
\end{equation}
Combining Eq. \ref{l1} and Eq. \ref{l2}, we obtain the following formulation:
\begin{equation}\label{option1}
\begin{aligned}
\mathcal{L}(W,\sigma_t,\sigma_m,\sigma_s,\sigma_d)
=\frac{1}{2\sigma^2_s}(\frac{1}{2\sigma_t^2}\mathcal{L}^{stu}_{traj}+\frac{1}{2\sigma_m^2}\mathcal{L}^{stu}_{man})\\
+\frac{1}{2\sigma^2_d}(\frac{1}{2\sigma_t^2}\mathcal{L}^{dis}_{traj}+\frac{1}{2\sigma_m^2}\mathcal{L}^{dis}_{man})
+F(\sigma_t,\sigma_m,\sigma_s,\sigma_d),\\
\end{aligned}
\end{equation}
where $F$ equals to $log\sigma_t\sigma_m(\frac{1}{2\sigma^2_s}+\frac{1}{2\sigma^2_d})
+log\sigma_s\sigma_d$. To ensure uniformity in the derived equations from different level-segmenting approach, we modify F as $F=log(\sigma_t\sigma_m\sigma_s\sigma_d)$.

\begin{table}[t]
  \centering
     \caption{Evaluation results for our proposed model and the other SOTA baselines in the NGSIM, HighD, and MoCAD datasets. \textbf{Bold} and \underline{underlined} values represent the best and second-best performance in each category. ``AVG'' is the average value of the RMSE.}\label{Table1}
     \vspace{1.5em}
     \setlength{\tabcolsep}{2mm}
   \resizebox{\linewidth}{!}{
    \begin{tabular}{c|ccccccc}
    \toprule
    \multicolumn{1}{c}{\multirow{2}[2]{*}{Dataset}} & \multirow{2}[3]{*}{Model} & \multicolumn{6}{c}{Prediction Horizon (s)} \\
\cmidrule{3-8}    \multicolumn{1}{c}{} &       & 1     & 2     & 3     & 4     & 5 & AVG\\
      \midrule
    \multirow{12}[20]{*}{NGSIM} 
    & S-LSTM   \cite{alahi2016social} & 0.65  & 1.31  & 2.16  & 3.25  & 4.55 & 2.38 \\
    & S-GAN    \cite{gupta2018social} & 0.57  & 1.32  & 2.22  & 3.26  & 4.40 & 2.35 \\
    & CS-LSTM  \cite{deo2018convolutional} & 0.61  & 1.27  & 2.09  & 3.10  & 4.37 & 2.29 \\
    & MATF-GAN \cite{zhao2019multi} & 0.66  & 1.34  & 2.08  & 2.97  & 4.13 & 2.22 \\
    & NLS-LSTM \cite{8813829} & 0.56  & 1.22  & 2.02  & 3.03  & 4.30 & 2.23 \\
    & IMM-KF   \cite{lefkopoulos2020interaction} & 0.58 & 1.36 & 2.28  & 3.37 & 4.55 & 2.43 \\
    & MFP      \cite{tang2019multiple} & 0.54  & 1.16  & 1.89  & 2.75  & 3.78 & 2.02 \\
    & DRBP     \cite{gao2023dual} & 1.18  & 2.83  & 4.22  & 5.82  & -  & 3.51 \\
    & WSiP     \cite{wang2023wsip} & 0.56  & 1.23  & 2.05  & 3.08  & 4.34 & 2.25 \\
    & CF-LSTM  \cite{Xu_Yang_Du_2020} & 0.55  & 1.10  & 1.78  & 2.73  & 3.82  & 1.99 \\
    & MHA-LSTM \cite{messaoud2021attention} & 0.41  & 1.01  & 1.74  & 2.67  & 3.83 & 1.91 \\
    & STDAN    \cite{chen2022intention} & \textbf{0.39} & \textbf{0.96} & 1.61 & 2.56 & 3.67 & 1.84 \\ 
    & iNATran  \cite{chen2022vehicle} & \underline{0.39} & \underline{0.96} & \underline{1.61} & 2.42 & 3.43  & \underline{1.76}  \\  
    & DACR-AMTP \cite{cong2023dacr}& 0.57  & 1.07 & 1.68 & 2.53 & 3.40 & 1.85  \\ 
    & FHIF   \cite{zuo2023trajectory} &{0.40} & {0.98} & 1.66 & 2.52 & 3.63 & 1.84  \\ 
     &BAT \cite{liao2024bat}& \textbf{0.23} & \textbf{0.81}  & {1.54}  & 2.52 &3.62 &1.74\\
    & HLTP++ & 0.46 & \underline{0.98} & \textbf{1.52} & \textbf{2.17} & \textbf{3.02} & \textbf{1.63}\\ 
    & HLTP++ (h) & 0.49 & 1.05 & 1.64 & \underline{2.34} & \underline{3.27} & \underline{1.76} \\
      \midrule
    \multirow{7}[25]{*}{HighD} &S-LSTM \cite{alahi2016social}& 0.22  & 0.62  & 1.27  & 2.15  & 3.41 &1.53 \\
    &S-GAN \cite{gupta2018social}& 0.30  & 0.78  & 1.46  & 2.34  & 3.41  &1.69 \\
    &WSiP \cite{wang2023wsip}& 0.20  & 0.60  & 1.21  & 2.07  & 3.14 &1.44 \\
    &CS-LSTM \cite{deo2018convolutional}& 0.22  & 0.61  & 1.24  & 2.10  & 3.27 &1.48 \\
    &MHA-LSTM \cite{messaoud2021attention}& 0.19  & 0.55  & 1.10  & 1.84  & 2.78 &1.29 \\
    &NLS-LSTM \cite{8813829}& 0.20  & 0.57  & 1.14  & 1.90  & 2.91 &1.34\\
    &DRBP\cite{gao2023dual}& 0.41  & 0.79  & 1.11  & 1.40  & -  & 0.92\\
    &EA-Net \cite{cai2021environment} & 0.15  & 0.26  & 0.43  & 0.78  & 1.32  &0.59 \\
    &CF-LSTM \cite{Xu_Yang_Du_2020}& 0.18  & 0.42  & 1.07  & 1.72  & 2.44 &1.17  \\
    &STDAN \cite{chen2022intention}& 0.19  & 0.27  & 0.48  & 0.91  & 1.66  &0.70 \\
    &DACR-AMTP \cite{cong2023dacr}& \textbf{0.10}  & \textbf{0.17}  & \underline{0.31}  & {0.54}  & {1.01} &0.42 \\
    &GaVa \cite{liao2024human}& 0.17  & 0.24  & 0.42  & 0.86  & 1.31  &0.60\\ 
    &HLTP++ & 0.11 & \textbf{0.17} & \textbf{0.30} & \textbf{0.47} & \textbf{0.75} & \textbf{0.36}\\
    &HLTP++ (h) & 0.12 & \underline{0.18} & 0.32 & \underline{0.52} & \underline{0.89} & \underline{0.41} \\
     \midrule
    \multirow{7}[11]{*}{MoCAD}
    &S-LSTM \cite{alahi2016social} & 1.73  & 2.46  & 3.39  & 4.01  & 4.93 &3.30 \\
    &S-GAN \cite{gupta2018social} & 1.69  & 2.25  & 3.30  & 3.89  & 4.69 &3.16 \\
    &CS-LSTM \cite{deo2018convolutional} & 1.45  & 1.98  & 2.94  & 3.56  & 4.49 &2.88 \\
    &MHA-LSTM \cite{messaoud2021attention} & 1.25  & 1.48  & 2.57  & 3.22  & 4.20  & 2.54\\
    &NLS-LSTM \cite{8813829} & 0.96  & 1.27  & 2.08  & 2.86  & 3.93 &2.22 \\
    &WSiP \cite{wang2023wsip} & 0.70  & 0.87  & 1.70  & 2.56  & 3.47 &1.86  \\
    &CF-LSTM \cite{Xu_Yang_Du_2020} & 0.72  & 0.91  & 1.73  & 2.59  & 3.44 &1.87 \\
    &STDAN \cite{chen2022intention} & \underline{0.62}  & \underline{0.85}  & \underline{1.62}  & \underline{2.51}  & 3.32 &1.78  \\
    &HLTP++  & \textbf{0.60} & \textbf{0.81} & \textbf{1.56} & \textbf{2.40} & \textbf{3.19}&\textbf{1.71} \\
    &HLTP++ (h) & 0.64 & 0.86 & \underline{1.62} & 2.52 & 3.35 & 1.80 \\
    \bottomrule
    \end{tabular}%
  }
\end{table}%

\section{Experiment}\label{Experiment}
\subsection{Experimental Setup}
\textbf{Datasets.} We conduct experiments using three esteemed datasets: NGSIM \cite{deo2018convolutional}, HighD \cite{8569552}, and MoCAD \cite{liao2024bat}. These three datasets cover various traffic conditions, including on highways and in dense urban environments. \\
\textbf{Metric.} Root Mean Square Error (RMSE) and average RMSE is applied as our primary evaluation metric, which is commonly used as a measure to calculate the square root of the average squared prediction error in autonomous driving.\\
\textbf{Implementation Details.}\label{Training} 
HLTP++ is developed using PyTorch and trained on an A40 48G GPU. We use the Adam optimizer along with CosineAnnealingWarmRestarts for scheduling, with a training batch size of 256 and learning rates ranging from $10^{-3}$ to $10^{-5}$. Unless specified, all evaluation results are based on the ``student'' model of HLTP++. 

\subsection{Experiment Results}
\textbf{Comparison with the State-of-the-art (SOTA) Baselines.}
Our comprehensive evaluation demonstrates HLTP++'s superior performance compared to state-of-the-art baselines, as detailed in Table \ref{Table1}. It notably achieves gains of 11.2\% for long-term (5s) and 11.4\% for average predictions on the NGSIM dataset. The corresponding outstanding performance is also evident on the HighD dataset and the MACAD dataset. It is noteworthy that our model HLTP++(h), despite utilizing only 1.5 seconds of input data (half of the input of other baselines), achieves comparable prediction accuracy. This highlights the adaptability and robustness of HLTP++.\\
\begin{table}[t]
  \centering
  \caption{Comparative evaluation of our model with SOTA baselines. Emphasizing model complexity via parameter count (Param.). ``Avg. IT'' represents the average inference time of the model. HLTP++(TM) is the teacher model of the HLTP++. HLTP++(SM) is the student model of the HLTP++ without KDM.}
  \vspace{1.5em}
    \setlength{\tabcolsep}{2mm}
    \resizebox{\linewidth}{!}{
    \begin{tabular}{c|ccccc}
    \toprule
    \multicolumn{1}{c}{\multirow{2}[2]{*}{Model}} & \multirow{2}[2]{*}{Param. (K)} & \multicolumn{3}{c}{Average RMSE (m)} & \multirow{2}[2]{*}{Avg. IT (s)} \\
\cmidrule{3-5}  \multicolumn{1}{c}{} &   & NGSIM & HighD & MoCAD & \\
    \midrule 
        CS-LSTM      & \underline{194.92} & 2.29 & 1.49 & 2.88 & \underline{0.0259} \\
        CF-LSTM      & 387.10 & 1.99 & 1.17 & 1.88 & 0.4565 \\
        WSiP         & 300.76 & 2.25 & 1.44 & 1.86 & 0.3292 \\
        STDAN        & 486.82 & 1.87 & 0.70 & 1.78 & 0.0670\\
        HLTP++       & \textbf{129.60} & \textbf{1.63} & \textbf{0.36} & \textbf{1.62} & \textbf{0.0214} \\
        HLTP++(SM)      & \textbf{129.60} & 1.76 & 0.52 & 1.75 & \textbf{0.0214} \\
        HLTP++(TM)      & 453.45 & \underline{1.74} & \underline{0.47} & \underline{1.72} & 0.0726\\
    \bottomrule
    \end{tabular}
    }
  \label{tab:parameters}%
\end{table}%
\textbf{Comparing Model Performance and Complexity.} 
As detailed in Table \ref{tab:parameters}, our benchmarking against SOTA baselines reveals that HLTP++ models outperform in all metrics while maintaining a minimal parameter count. Specifically, HLTP++ reduce parameters by 56.91\% and 33.51\% compared to WSiP and CS-LSTM, respectively. Compared to HLTP++(SM), the ``teacher'' model of HLTP++, HLTP++(TM), achieve the second best score in three datasets, while maintaining a larger number of parameters and slower inference speed. However, HLTP++ maintain the lowest inference time while achieve the best accuracy in trajectory prediction. Utilizing the Knowledge Distillation Module (KDM), HLTP++ retains the lightweight advantages of the HLTP++(SM), while concurrently enhancing its predictive capabilities by assimilating knowledge gleaned from the teacher model, thereby surpassing the performance of the teacher model itself. This highlights the efficiency and adaptability of our lightweight ``teacher-student'' knowledge distillation framework, offering a balance between accuracy and computational resources.\\
\textbf{Qualitative Results.}
Figure \ref{pic-compare} showcase the multimodal probabilistic prediction performance of HLTP++ on the NGSIM dataset. The heat maps shown represent the Gaussian Mixture Model of predictions in challenging scenes. These visualizations show that the highest probability predictions of our model are very close to the ground truth, indicating its impressive performance. Figure \ref{pic-multi1} visually demonstrates our model's ability to accurately predict complex scenarios such as merging and lane changing, confirming its effectiveness and safety in various traffic situations. Interestingly, in certain complex scenarios, the trajectory predictions of the ``student'' model exceed the accuracy of the ``teacher'' model. This result illustrates the ability of the ``student'' model to selectively assimilate and refine the knowledge acquired from the ``teacher'' model, effectively ``extracting the essence and discarding the dross''. 
Moreover, we observed that vehicles in closer proximity to the target vehicle received higher attention values. This observation aligns with the driving behavior of human operators who primarily focus on the vehicle ahead, as it has the most significant influence on the driving trajectory. This also substantiates the utility of vision pooling in reducing the perturbations caused by neighboring vehicles, thereby prioritizing the significance of the leading vehicle.
\begin{table}[t]
    \vspace{1em}
  \centering
        \caption{Different methods and components of ablation study.}\label{(1)}
        \vspace{1em}
        \setlength{\tabcolsep}{3mm}
        \resizebox{\linewidth}{!}{
            \begin{tabular}{cccccccc}
                \toprule
                \multirow{2}[2]{*}{Components} & \multicolumn{7}{c}{Ablation Methods} \\
                \cmidrule{2-8}          & A     & B     & C     & D     & E  &F &G \\
                \midrule
                Visual Pooling Mechanism & \ding{56} & \ding{52} & \ding{52} & \ding{52} & \ding{52} & \ding{52} & \ding{52} \\
                Spatial Encoder & \ding{52} & \ding{56} & \ding{52}& \ding{52}  & \ding{52} & \ding{52}  & \ding{52}\\
                Fusion Module & \ding{52} & \ding{52} &  \ding{56} &\ding{52} & \ding{52}  & \ding{52}  & \ding{52}\\
                FA-SNN & \ding{52} & \ding{52} & \ding{52} & \ding{56} & \ding{52}  & \ding{52}  & \ding{52}\\
                Multimodal Decoder & \ding{52} & \ding{52} & \ding{52} & \ding{52}& \ding{56}  & \ding{52}  & \ding{52}\\
                KDM & \ding{52} & \ding{52} & \ding{52} & \ding{52}& \ding{52}  & \ding{56}  & \ding{52}\\
                \bottomrule
            \end{tabular}}%
\end{table}

\begin{table}[t]
      \centering
          \caption{Ablation results for different models. (NGSIM/HighD)}
          \label{(2)}
          \vspace{1em}
           \setlength{\tabcolsep}{2mm}
          \resizebox{\linewidth}{!}{
            \begin{tabular}{cccccccc}
            \toprule
            \multirow{2}[2]{*}{Time (s)} & \multicolumn{6}{c}{Ablation Methods} \\
        \cmidrule{2-7} & A & B & C & D & E & F \\
            \midrule
             1 & 0.52/0.11 & 0.47/0.11 & 0.53/0.12 & 0.47/0.11 & 0.49/0.20 & 0.50/0.16 \\
             2 & 1.07/0.18 & 1.03/0.20 & 1.09/0.23 & 1.10/0.17 & 1.10/0.26 & 1.06/0.32 \\
             3 & 1.68/0.33 & 1.71/0.34 & 1.74/0.34 & 1.67/0.32 & 1.81/0.44 & 1.64/0.45 \\
             4 & 2.35/0.49 & 2.40/0.51 & 2.41/0.49 & 2.34/0.50 & 2.63/0.72 & 2.33/0.69 \\
             5 & 3.24/0.78 & 3.39/0.80 & 3.32/0.79 & 3.27/0.79 & 3.67/0.97 & 3.26/0.98 \\
             AVG & 1.77/0.38 & 1.80/0.39 & 1.82/0.39 & 1.77/0.38 & 1.94/0.52 & 1.76/0.52 \\
            \bottomrule
            \end{tabular}}%
\end{table}%

\begin{figure*}[htbp]
    \centering
    \includegraphics[width=0.92\linewidth]{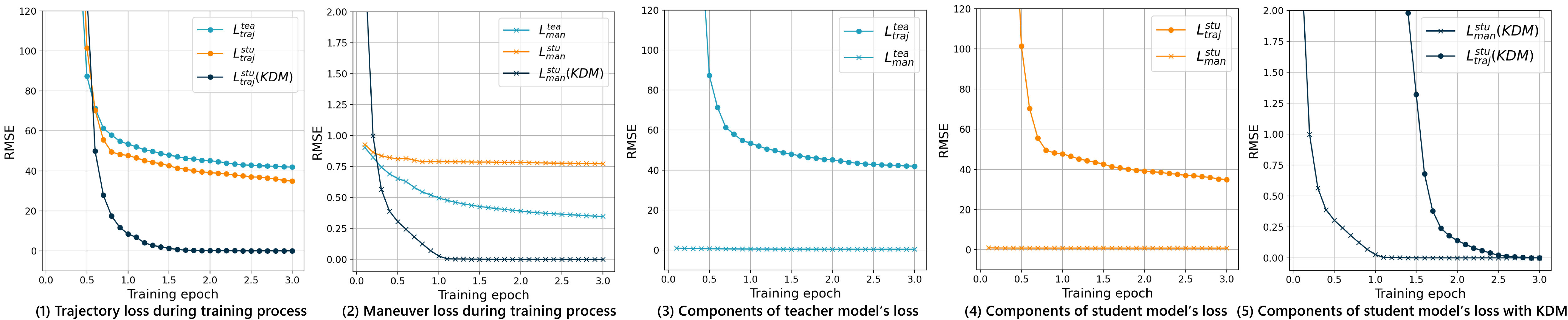}
    \caption{Visualizations of the (1) trajectory loss, (2) maneuver loss, (3) teacher model's loss, (4) student model's loss without KDM, (5) student model's loss with KDM during training.}
    \label{fig:kdm}
\end{figure*}

\begin{figure}[t]
    \centering
    \includegraphics[width=\linewidth]{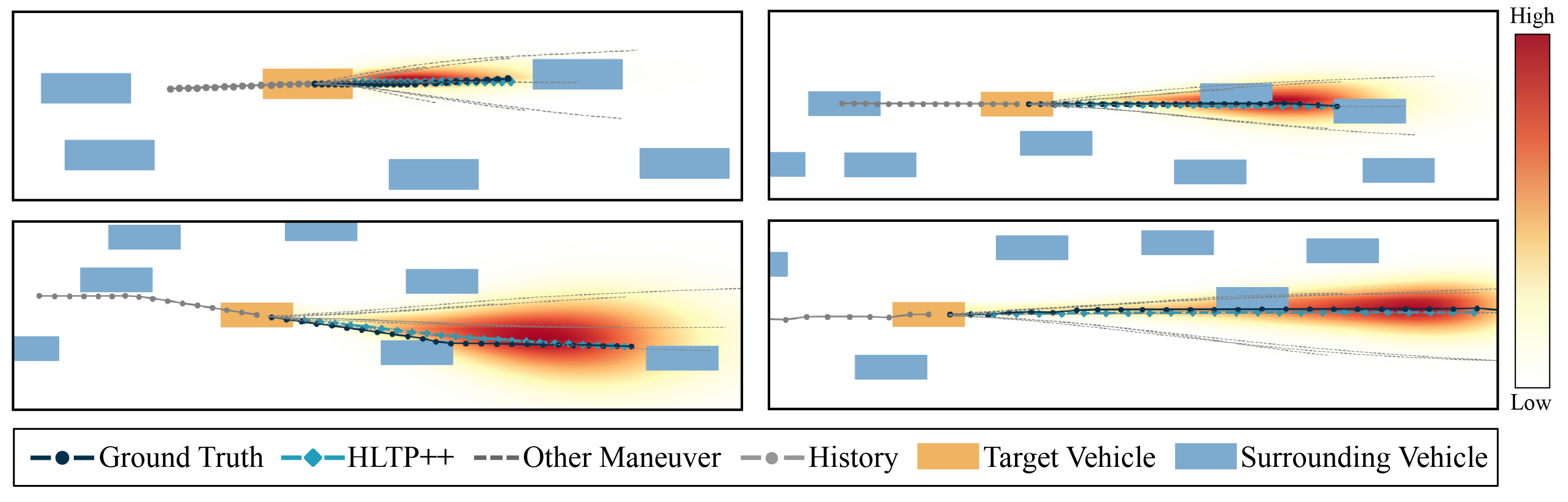}
    \caption{Visualizations of the multi-modal probabilistic prediction of HLTP++ on NGSIM. Heat maps illustrate the GMM of predictions: brighter areas denote higher probabilities.}
    \vspace{2em}
    \label{pic-compare}
\end{figure}

\begin{table}[t]
  \centering
        \caption{Ablation study on FA-SNN.}\label{fa-snn}
        \vspace{1em}
        \setlength{\tabcolsep}{5mm}
        \resizebox{0.85\linewidth}{!}{
            \begin{tabular}{ccccc}
               \bottomrule
                \multicolumn{1}{c}{\multirow{2}[2]{*}{Models}} & \multicolumn{2}{c}{Components} & \multicolumn{2}{c}{Metrics}\\
                \cmidrule(lr){2-3} \cmidrule(lr){4-5} & AST & FT & RMSE (5s) & AVG \\
                \midrule
                FA-SNN & \ding{52} & \ding{52} & 3.02 & 1.63 \\
                F-SNN  & \ding{56} & \ding{52} & 3.13 & 1.70 \\
                A-SNN  & \ding{52} & \ding{56} & 3.11 & 1.69 \\
                SNN    & \ding{56} & \ding{56} & 3.21 & 1.73 \\
                \toprule 
            \end{tabular}}%
\end{table}
\subsection{Ablation Studies}

\textbf{Ablation Study for Core Components.} 
Table \ref{(1)} shows that our ablation study evaluates the performance of HLTP++ using six model variations, each omitting different components. 
The data in Table \ref{(2)} clearly indicate that the performance of all models degrades when components are removed, as compared to the baseline model. Notably, integrating the iTransformer and a multimodal probabilistic maneuvering module significantly improves the accuracy. This underscores their vital function in encapsulating the spatio-temporal dynamics among vehicles. Furthermore, ablation studies A and B provide empirical evidence supporting the utility of incorporating cognitive mechanisms similar to human brain processes.\\

\textbf{Ablation Study for FA-SNN.}
To further showcase the effectiveness of our proposed FA-SNN, we conducted an ablation study in HLTP++ by replacing the FA-SNN with the standard SNN, SNN only with the Fourier Transform (FT), and the SNN only with the adaptive spike threshold (AST). Table \ref{fa-snn} demonstrates that incorporating the AST and FT in SNN can significantly improve the predictive performance of the model. This underscores the FA-SNN's ability to capture and extract spatio-temporal interactions in complex scenes.\\
\textbf{Ablation Study for KDM.} 

To better illustrate the impact of KDM on HLTP++, we present the curves of different loss functions during the training process, as shown in Figure \ref{fig:kdm} (1) and \ref{fig:kdm} (2).  Specifically,  $L^{tea}_{traj}$, $L^{stu}_{traj}$, $\widetilde{L}^{stu}_{traj}$(KDM), and $L^{stu}_{traj}$(KDM) represent the trajectory losses of training the teacher model independently, training the student model independently, training the student model with KDM, respectively. Accordingly, $L^{tea}_{man}$, $L^{stu}_{man}$, and $L^{stu}_{man}$(KDM) denote the maneuver losses.
From Figure \ref{fig:kdm} it can be seen that both $L^{stu}_{traj}$(KDM) and $L^{stu}_{man}$(KDM) have initially higher values during training, but they quickly exhibit exponential decay as training progresses. This ensures a more efficient initial descent of the model during training and allows for adjustments at a finer granularity in later stages, where the traditional loss functions fail to achieve. 

Furthermore, in Figure \ref{fig:kdm} (3)-(5), we compare the trajectory loss and maneuver loss between the teacher model, the student model without KDM, and the student model with KDM. In Figure \ref{fig:kdm} (3) and Figure \ref{fig:kdm} (4), there is a significant difference in the loss values between the instructor and student models, and as training progresses, the difference between them stabilizes around 40, differing by two orders of magnitude. Even in these two subplots, it is difficult to observe a noticeable decrease in maneuver loss. This implies that without KDM adjustments, there would be a severe imbalance in the ratio between trajectory loss and maneuver loss, causing the model to lean more towards trajectory fitting while neglecting lane change information.
However, Figure \ref{fig:kdm} (5) shows that although the magnitudes of $L^{stu}_{traj}$(KDM) and $L^{stu}_{man}$(KDM) are significantly different at the beginning of training, they quickly converge as training progresses. This indicates that adaptive tuning through KDM allows for a better balance of the proportions between different loss, thus achieving a balanced optimization of loss functions for different tasks.\\
\textbf{Ablation Study for Missing Data.} We introduce a missing test set on the NGSIM dataset, focusing on scenarios where part of the historical data is missing. The set is divided into five subsets based on varying durations of data absence.
For example, subset $t_m$=0.4 indicates a missing trajectory data duration of 0.4 seconds, which was imputed using linear interpolation. The results in Table \ref{missing-data} show that HLTP++ outperforms all baselines even with 1.6s missing data, highlighting its adaptability and deep understanding of traffic dynamics. 

\begin{figure}[htbp]
\centering
\includegraphics[width=\linewidth]{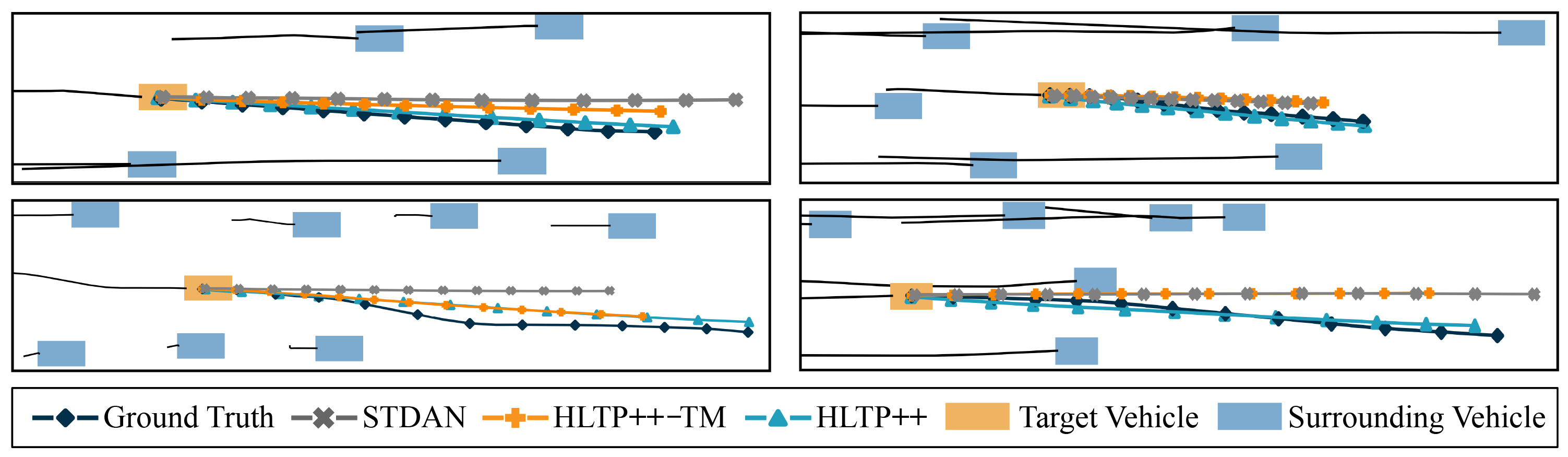}
\caption{Visualizations of HLTP++ and SOTA baseline on NGSIM. HLTP++(TM) denotes the ``teacher'' model of HLTP++.}\label{pic-multi1}
\vspace{2em}
\end{figure}

\begin{table}[htbp]
  \centering
    \caption{Ablation results for missing data.}
    \vspace{1em}
    \label{missing-data}
        \setlength{\tabcolsep}{2mm}
        \resizebox{0.85\linewidth}{!}{
        \begin{tabularx}{\linewidth}{c|ccccccc}
        \toprule
        \multicolumn{1}{c}{Time (s)}& $t_m$=0.4 & $t_m$=0.8 & $t_m$=1.2 & $t_m$=1.6 & $t_m$=2.0 & $t_m$=2.4 \\
        \midrule
                1 & 0.46 & 0.46 & 0.47 & 0.48 & 0.50 & 0.56\\ 
                2 & 0.98 & 0.99 & 1.00 & 1.03 & 1.07 & 1.17\\ 
                3 & 1.53 & 1.54 & 1.57 & 1.61 & 1.67 & 1.79\\ 
                4 & 2.20 & 2.21 & 2.25 & 2.31 & 2.39 & 2.54\\ 
                5 & 3.07 & 3.10 & 3.15 & 3.23 & 3.32 & 3.51\\
        \bottomrule
        \end{tabularx}%
        }
        \vspace{1em}
\end{table}%
\vspace{-1em}
\section{Conclusion}\label{Conclusion}
This study presents a novel trajectory prediction model (HLTP++) for AVs. It addresses the limitations of previous models in terms of parameter heaviness and applicability. HLTP++ is based on a multi-level task knowledge distillation network, providing a lightweight yet efficient framework that maintains prediction accuracy. Importantly, HLTP++ adapts to scenarios with missing data and reduced inputs by simulating human observation and making human-like predictions. The empirical results indicate that HLTP++ excels in complex traffic scenarios and achieves SOTA performance. In future work, we plan to feed multimodal data, such as Bird's Eye View (BEV), multi-view camera images and Lidar, into the HLTP++ model to further enhance the scene understanding of the model.

\section*{Acknowledgements}
This research is supported by the Science and Technology Development Fund of Macau SAR (File no. 0021/2022/ITP, 0081/2022/A2, 001/2024/SKL), Shenzhen-Hong Kong-Macau Science and Technology Program Category C (SGDX20230821095159012), and University of Macau (SRG2023-00037-IOTSC). Haicheng Liao and Yongkang Li contributed equally to this work. Please ask Dr. Zhenning Li (zhenningli@um.edu.mo) for correspondence.



\bibliography{mybibfile}
\end{document}